\documentclass[pmlr,twocolumn,10pt]{jmlr} 

\usepackage{booktabs}
 
\usepackage{siunitx}

\usepackage[switch]{lineno}

\usepackage{multirow}
\usepackage{makecell}
\usepackage{amssymb}
\usepackage{array}
\usepackage{cleveref}
\usepackage{breakcites} 
\usepackage{pifont}
\usepackage{subcaption}

\definecolor{meshcolor1}{HTML}{d81b60}
\definecolor{meshcolor2}{HTML}{1e88e5}

\newcommand{\cmark}{\ding{51}}%
\newcommand{\xmark}{\ding{55}}%

\newcommand{\qed}{\hfill\ensuremath{\blacksquare}}

\theorembodyfont{\upshape}
\theoremheaderfont{\scshape}
\theorempostheader{:}
\theoremsep{\newline}

\jmlrvolume{259}
\jmlryear{2024}
\jmlrsubmitted{}
\jmlrpublished{}
\jmlrworkshop{Machine Learning for Health (ML4H) 2024} 

 \title[MLV\textsuperscript{2}-Net]{MLV\textsuperscript{2}-Net: Rater-Based \underline{M}ajority-\underline{L}abel \underline{V}oting for Consistent \underline{M}eningeal \underline{L}ymphatic \underline{V}essel Segmentation}

 \author{%
  \Name{Fabian Bongratz}\nametag{$^{a,b}$} \Email{fabi.bongratz@tum.de}\\
  \Name{Markus Karmann}\nametag{$^{a}$} \Email{markus.karmann@tum.de}\\
  \Name{Adrian Holz}\nametag{$^{a}$} \Email{adrian.holz@tum.de}\\
  \Name{Moritz Bonhoeffer}\nametag{$^{a}$} \Email{moritz.bonhoeffer@tum.de}\\
  \Name{Viktor Neumaier}\nametag{$^{a}$} \Email{viktor.neumaier@tum.de}\\
  \Name{Sarah Deli}\nametag{$^{a}$} \Email{sa.deli@tum.de}\\
  \Name{Benita Schmitz-Koep}\nametag{$^{a}$} \Email{benita.schmitz-koep@tum.de}\\
  \Name{Claus Zimmer}\nametag{$^{a}$} \Email{claus.zimmer@tum.de}\\
  \Name{Christian Sorg}\nametag{$^{a}$} \Email{christian.sorg@tum.de}\\
  \Name{Melissa Thalhammer}\nametag{$^{a}$} \Email{melissa.thalhammer@tum.de}\\
  \Name{Dennis M.~Hedderich}\nametag{$^{a}$} \Email{dennis.hedderich@tum.de}\\
  \Name{Christian Wachinger}\nametag{$^{a,b}$} \Email{christian.wachinger@tum.de}\\
  \addr $^{a}$School of Medicine and Health, Technical University of Munich, Munich, Germany\\
  \addr $^{b}$Munich Center for Machine Learning, Munich, Germany
 }

\begin{document}

\maketitle

\begin{abstract}
Meningeal lymphatic vessels (MLVs) are responsible for the drainage of waste products from the human brain. An impairment in their functionality 
has been associated with aging as well as brain disorders like multiple sclerosis and Alzheimer's disease. However, MLVs have only recently been 
described for the first time in magnetic resonance imaging (MRI), and their ramified structure renders manual segmentation particularly difficult. Further, as there is no consistent notion of their appearance, human-annotated MLV structures contain a high inter-rater variability that most automatic segmentation methods cannot take into account.
In this work, we propose a new rater-aware training scheme for the popular nnU-Net model, and we explore rater-based ensembling strategies for accurate and consistent segmentation of MLVs. This enables us to boost nnU-Net's performance while obtaining explicit predictions in different annotation styles and a rater-based uncertainty estimation. Our final model, \emph{MLV\textsuperscript{2}-Net}, achieves a Dice similarity coefficient of 0.806 with respect to the human reference standard. The model further matches the human inter-rater reliability and replicates age-related associations with MLV volume.
\end{abstract}
\begin{keywords}
Meningeal lymphatic vessels, Glymphatic system, Segmentation, Inter-rater variability
\end{keywords}

\paragraph*{Data and Code Availability}

In \Cref{tab:dataset_structure}, we provide an overview of the data used in this study. As no public data on MLV structures in MRI is available, we assembled a custom segmentation dataset comprising $n=33$ labeled and $n=22$ unlabeled 3D fluid-attenuated inversion recovery (FLAIR) magnetic resonance (MR) images from cognitively normal subjects. To this end, four neuroanatomical experts annotated MLV structures individually along the superior sagittal sinus (SSS) in anterior, middle, and posterior brain regions, resulting in $3{\times}7$ and $1{\times}6$ annotations per rater.
In addition, two images were annotated by all raters, which allows us to assess their inter-rater reliability (IRR). We keep another held-out test set comprising four more images to evaluate model accuracy. All raters annotated those images jointly for best consistency after the annotation of the $27 + 2$ scans was finished. 
Finally, we use 22 raw images to evaluate our model indirectly by replicating known age-related associations with MLV volume.
All images have a resolution of $0.5{\times}0.5{\times}1\text{mm}^3$ and were acquired using simultaneous trimodal PET-MR-EEG imaging~\citep{trimage}. We will make code and trained models publicly available at \url{https://github.com/ai-med/mlv2-net}.

\begin{figure*}[t]
    \centering
     \includegraphics[width=\textwidth]{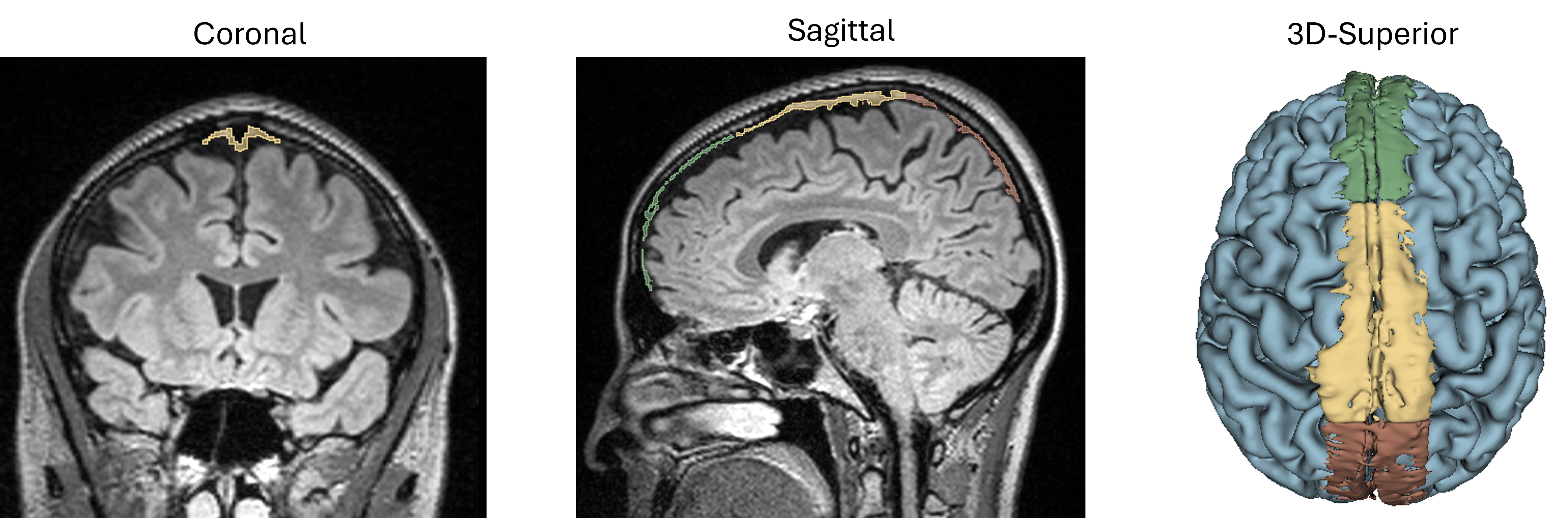}
    \caption{Exemplary segmentation of meningeal lymphatic vessels (MLVs) in coronal and sagittal planes --- separated into anterior, middle, and posterior regions. On the right, we show the corresponding 3D surface representation rendered from a superior viewpoint.}
    \label{fig:task}
\end{figure*}

\paragraph*{Institutional Review Board (IRB)}
Imaging data comes from two studies. Both studies were approved by the Ethics Review Board of the Klinikum Rechts der Isar, Technical University of Munich, Germany (5611/12, 338/20-S). Approvals to administer radiotracers for both studies were obtained from the Administration of Radioactive Substances (Bundesamt fuer Strahlenschutz), Germany (Z 5-22461/2 – 2014-010, Z 5- 22464/2020-199-G).

\section{Introduction}
The lymphatic system --- part of the immune system and responsible for the drainage of waste products --- stretches across the entire human body and can often be found alongside
blood vessels of the circulatory system. In the brain, the glymphatic system~\citep{Iliff2012} takes a similar role in that it clears waste products. To this end, meningeal lymphatic vessels (MLVs), located alongside the dural venous sinuses, transfer interstitial fluids and macromolecules to deep cervical lymph nodes~\citep{Louveau2015}. An impairment in the MLVs' functionality, potentially coupled with morphological changes such as thickening, has been linked to aging~\citep{albayram2022non} as well as to clinical conditions like Alzheimer's~\citep{Goodman2018}, multiple sclerosis~\citep{Louveau2018}, and Parkinson’s disease~\citep{Ding2021}. Yet, MLVs have only recently been described in 3D FLAIR MRI~\citep{albayram2022non}, and their segmentation has only been done manually so far. However, manual annotation of MLVs is difficult and time-consuming due to their ramified structure, cf.~\Cref{fig:task}. Moreover, the training of automatic segmentation models on expert-annotated data is challenging due to the high inter-rater variability.

\noindent
\paragraph{Related work}
Deep neural networks for medical image segmentation are commonly trained to remove this variablility~\citep{Guo2024,hatamizadeh2022unetr,ronnebergerUNetConvolutionalNetworks2015}. However, this approach does not model the reality where disagreement about the true contours of a structure often exists~\citep{Warfield2004}. This issue is especially problematic for newly discovered structures, such as MLVs, which bear enormous potential for innovative findings but for which a common notion of their appearance does not (yet) exist. 
Notably, a few dedicated methods for rater-aware segmentation were developed~\citep{kohl2018-probabilistic-unet,deep_learning_esnambles_from_multiple_annotations,Warfield2004,learning_from_multiple_annotators_for_medical_image_segmentation}. 
These approaches yielded effective results for certain standard applications, e.g., skin lesion~\citep{deep_learning_esnambles_from_multiple_annotations} or brain tumor segmentation~\citep{learning_from_multiple_annotators_for_medical_image_segmentation}, but transferring them to new tasks is difficult due to the large number of hyperparameters involved. These choices are non-trivial, not reproducible, and subject to the developer's experience and preferences~\citep{nnunet}. At the same time, the best segmentation results are typically obtained with nnU-Net~\citep{isensee2024nnunet-revisited}, which provides a versatile framework for hyperparameter selection.
Unfortunately, nnU-Net cannot model the variability in segmentations provided by different raters --- a functionality essential for trustworthy and comprehensible clinical predictions. We close this gap and develop a rater-based ensembling strategy for nnU-Net that keeps its architecture intact and augments it with the ability to replicate individual raters' annotation styles. 

\noindent
\paragraph{Contribution}
We present the first automatic method for segmentation of MLVs from 3D FLAIR MRI. To achieve accurate and reliable segmentation of the ramified structure, we made the following technical contributions. First, we developed an innovative rater-aware training scheme for the popular nnU-Net model that takes into account the different raters involved in the creation of the training set. This enables nnUNet to learn individual raters' segmentation styles and to explicitly predict a set of plausible segmentations. In a second step, we aggregate the predictions with a weighted majority-label voting scheme for best segmentation accuracy. In addition, we obtain a rater-based uncertainty prediction from the model. 
Finally, since the volume of MLVs is usually of utmost importance for downstream analyses, we derive error boundaries of the model's predicted volumes with respect to the ground-truth volume.

\begin{table*}[t]
    \centering
    \caption{Composition of the annotated and raw datasets used in this study. 
    } 
    \label{tab:dataset_structure}
        \begin{tabular}{l l c c c}
    	\hline
        \multirow{2}{*}{Name} & \multirow{2}{*}{Used for} & Joint annot{.} & \#Annotations & \multirow{2}{*}{\#Images}\\
    	& &  by all raters & per image  & \\
    	\hline
    	Training set & Training \& validation & \xmark & $1$ &  $27$\\
        IRR set & Testing (inter-rater reliability) & \xmark & $4$ & $2$\\
        Consensus set & Testing (accuracy) & \cmark & $1$ & $4$\\
        Raw set & Testing (downstream analysis) & N/A & N/A & $22$\\
    	\hline
    \end{tabular}
    
\end{table*}

\section{Methods}

\subsection{MLV$^2$-Net architecture}

\begin{figure*}[t]
  \centering
  \includegraphics[width=\textwidth]{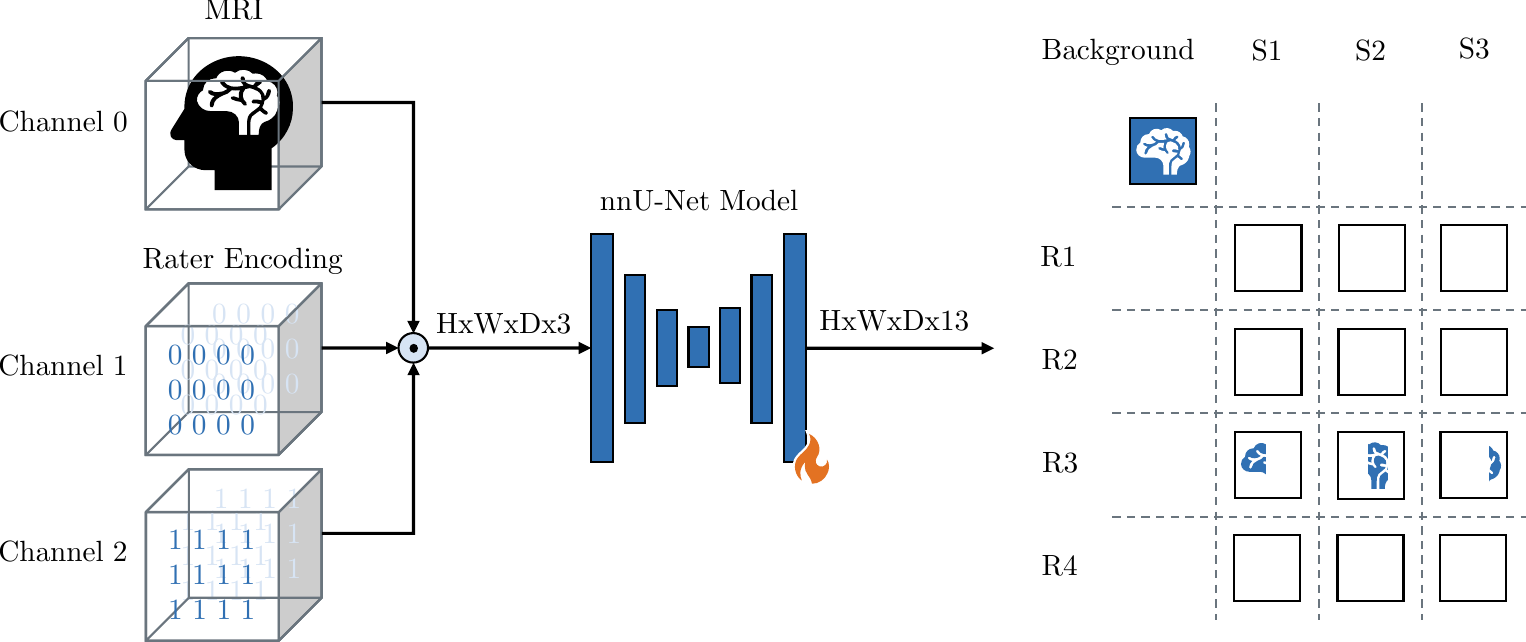}
  \caption{Illustration of MLV$^2$-Net. It augments nnU-Net with a rater-specific encoding and yields rater-aware segmentations as output. `$\odot$` denotes a channel-wise concatenation of inputs. In this example, we show the encoding and segmentation output figuratively for rater 3 (R3) and three foreground segmentation labels ($S1-S3$). We train nnU-Net from scratch, as indicated by the flame.} \label{fig:annotatorinputconcept}
\end{figure*}

\Cref{fig:annotatorinputconcept} shows an overview of MLV$^2$-Net, which stands for rater-based \underline{m}ajority-\underline{l}abel-\underline{v}oting-network for \underline{m}eningeal \underline{l}ymphatic \underline{v}essels. MLV$^2$-Net builds upon nnU-Net~\citep{nnunet} and takes a 3D image of shape $H{\times}W{\times}D$ as input. In addition, we incorporate a unique encoding of the rater (rater encoding) of the same shape as the image. The rationale behind this encoding is that it provides relevant information about the rater, in our case a neuroanatomical expert who provides annotations of MLVs, without any architectural changes that might derange nnU-Net's hyperparameter search strategy. As output, the network yields voxel-wise segmentation maps, separated by foreground class and rater.
Eventually, all predictions are aggregated via weighted majority-label voting as shown in \Cref{fig:majority-label-vote}. 
Apart from the input and output, we keep nnU-Net intact; hence, we benefit from its structured parameter selection and obtain a reproducible setup. 

\subsection{Rater-aware training and inference}
During the training and inference of MLV$^2$-Net, we consider the different raters in the input and the output of the model. 

\noindent
\paragraph{Rater as input}
To enable the network to learn the styles of different raters from the training data, we provide this information as input to the network. Technically, we assign a zero-centered one-hot-encoded ID to each rater and concatenate it as additional channels to the input image volume as depicted in \Cref{fig:annotatorinputconcept,fig:majority-label-vote}. 
Namely, we assign the four raters in our setting the codes [1,~0], [-1,~0], [0,~1], and [0, -1].
In general, this scheme leads to $R/2$ additional input channels for $R$ raters, which can be well processed by nnU-Net's initial convolutional layer. Importantly, we disable the per-channel z-score normalization in nnU-Net for the rater-encoding channels, which would set them to zero and essentially erase the rater information.

\noindent
\paragraph{Rater as output}
To enforce the model to consider the rater, i.e., to coerce the network to predict the correct structure \emph{as annotated by a certain rater},
we create rater-specific foreground labels for the loss computation (a combination of cross-entropy and Dice loss as in nnUNet). These are also the labels predicted by MLV$^2$-Net during inference. Specifically, we use labels ``Anterior MLV/Rater~1'', ``Anterior MLV/Rater~2'', ``Anterior MLV/Rater~3'', and ``Anterior MLV/Rater~4'' instead of one label ``Anterior MLV''. Likewise for labels ``Middle MLV'' and ``Posterior MLV''. This results in $RF + 1$ labels, i.e., the cartesian product of $R$ rater and $F$ foreground labels, and the background. We do not create rater-specific background labels, as this would be redundant.

\subsection{Weighted majority-label voting}\label{sec:nnunet_combining_annotator_specific_predictions}

\begin{figure*}[t]
  \centering
  \includegraphics[width=0.8\textwidth]{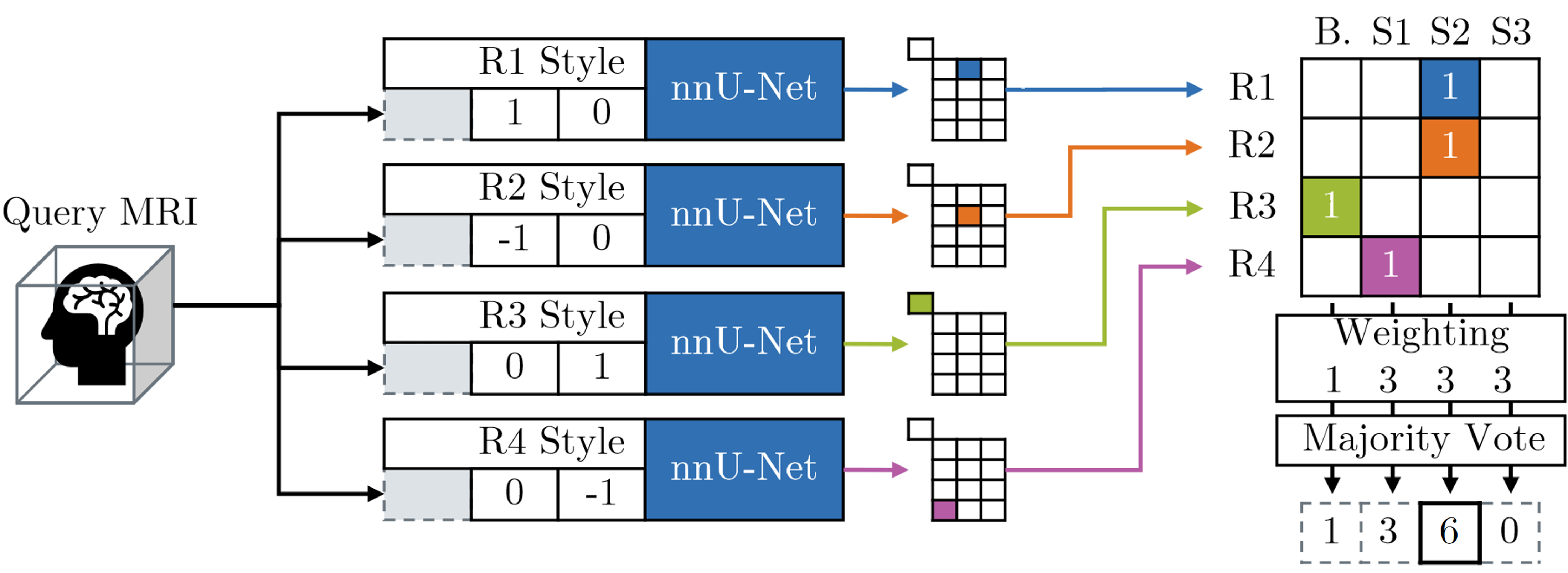}
  \caption{Illustration of the weighted majority-label voting in MLV$^2$-Net for four raters (R1--R4). Shown is an exemplary ambiguous prediction for a single voxel. R3 would segment the voxel as background (B), R4 as foreground segment 1 (S1), and R1 and R2 as foreground segment 2 (S2). The foreground weight in the example is set to $w_{\text{fg}}=3$.
  }
  \label{fig:majority-label-vote}
\end{figure*}

To aggregate the segmentation maps in the styles of different raters, we employ a weighted majority-label voting, which we illustrate in
\Cref{fig:majority-label-vote}.
As shown, we multiply the number of votes from rater-specific foreground predictions by a weight $w_{\text{fg}} > 1$. Compared to the standard majority vote ($w_{\text{fg}}=1$), this increases the sensitivity to foreground labels, which we found to correspond best to human consensus decision-making (cf.~\Cref{sec:consens}). In the rare case of a tie, we choose the class with the lowest index. By default, nnU-Net also creates an ensemble of models via cross-validation and aggregates the voxel-wise mean of the predicted logits. We keep this mechanism untouched, i.e., cross-validation ensembling is part of the nnU-Net blocks in \Cref{fig:majority-label-vote}, and compute the majority-label vote externally, treating each cross-validation ensemble as one voting model.

\subsection{Rater-based uncertainty}
Apart from the consensus prediction, we obtain a rater-based uncertainty map in MLV$^2$-Net. The uncertainty is based on the agreement of rater-based predictions of the model, i.e., the uncertainty is higher the more rater-based predictions speak against the majority label for a certain voxel. This provides us with an estimate of the reliability of the prediction, which renders the model more faithful as it can be used to detect potential failure cases. Unlike alternative uncertainty estimation methods, e.g., Monte Carlo Dropout~\citep{gal16dropout}, MLV$^2$-Net does not require a variational network architecture but keeps nnU-Net overall intact. Another advantage of our explicit rater-based modeling is that individual, potentially flawed, or deprecated segmentation styles can easily be ignored post hoc, i.e., without re-training. This is typically impossible with variational approaches that implicitly model the data variability.

\subsection{Boundaries on segmented volume based on Dice}
The performance of segmentation models is commonly evaluated with the Dice similarity coefficient ($DSC$). However, in the end, the segmented volume is often of utmost importance in medical imaging. Therefore, in the following, we derieve error boundaries on the predicted volume relative to the ground-truth or reference volume.

\begin{theorem}
    Given the Dice similarity coefficient ($DSC$) of a segmentation model, the predicted volume relative to the ground-truth volume is bounded by $\frac{2}{DSC} - 1$ from above and by $\frac{2}{2 - DSC} - 1$ from below.
\end{theorem}
\noindent 
\textit{Proof.} The $DSC$ and the relative predicted volume $V^{\mathrm{rel}}$ can be calculated from a confusion matrix comprising false negative ($FN$), true positive ($TP$), false positive ($FP$), and true negative ($TN$) voxels. By definition, the $DSC$ is calculated as
\begin{equation}
    DSC = \frac{2 \cdot TP}{2 \cdot TP + FN + FP} = \frac{2 \cdot TP}{1 + TP + FP},
    \label{eqn:dice_from_conf}
\end{equation}
where we utilized the fact that $TP + FN \equiv 1$ when normalized to the ground-truth (GT) volume.
The relative predicted volume, i.e., the predicted relative to the GT volume, is given by
\begin{equation}
    V^{\text{rel}} = \frac{\text{Predicted volume}}{\text{GT volume}} = \frac{TP + FP}{TP + FN} = TP + FP.
    \label{eqn:pred_volume_from_conf}
\end{equation}
Rearranging \Cref{eqn:dice_from_conf} to $FP = \frac{2 \cdot TP}{DSC} - TP - 1$ and inserting it into \Cref{eqn:pred_volume_from_conf} yields
\begin{equation}
    V^{\text{rel}} = \frac{2 \cdot TP}{DSC} - 1. 
    \label{eqn:Vrel_based_on_tp_and_dice}
\end{equation}
From $TP\leq1$, we obtain $V^{\text{rel}} \leq \frac{2}{DSC} - 1$. Similarly, we get $V^{\text{rel}}\geq \frac{2}{2 - DSC} - 1$ by rearranging \Cref{eqn:dice_from_conf} to $TP = \frac{DSC \cdot (FP + 1)}{2 - DSC}$ and $FP\geq 0$. \qed

\section{Results}

\begin{figure*}[t]
  \centering
  \includegraphics[width=\textwidth]{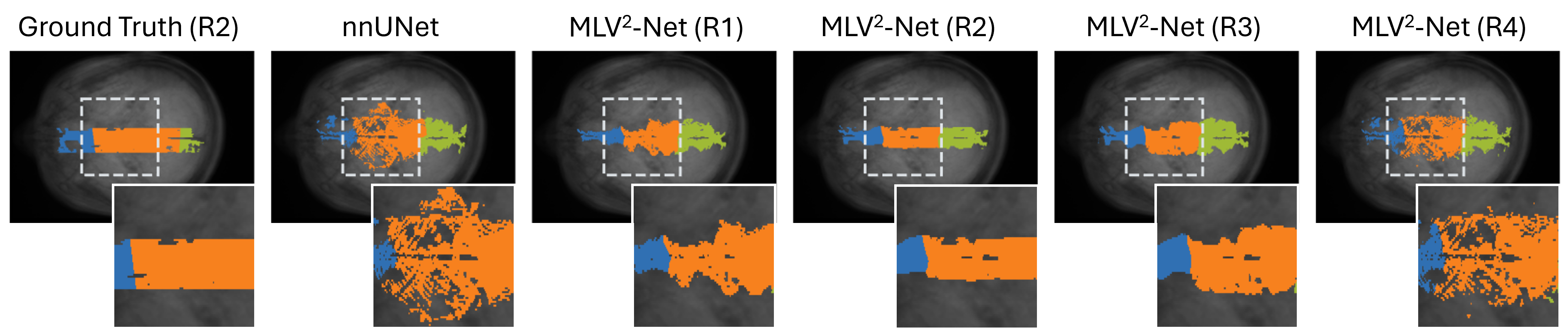}
  \caption{Ground-truth annotation from rater 2 (R2) and corresponding predictions from nnUNet and MLV$^2$-Net with different rater queries (R1--R4). We show projections along the vertical axis for best visibility; note that this makes the vessels appear thicker but better displays segmentation characteristics than slices.}
  \label{fig:individual-preds}
\end{figure*}

\begin{table*}[t]
\setlength{\tabcolsep}{3.5pt}
    \centering
    \caption{Accuracy of all implemented methods in terms of Dice similarity coefficient ($DSC$). We report results (mean $\pm$ SD) from the cross-validation and the held-out consensus test set in anterior, middle, and posterior regions and foreground (all regions combined). \textbf{Best} and $\underline{\text{second}}$ results are highlighted.}
    \label{tab:accuracy}
        {\renewcommand\bfdefault{b}
\begin{tabular}{l l c c c c }
	\toprule
    & & \multicolumn{3}{c}{Cross-validation} & Consensus set \\
    \cmidrule(lr){3-5}
    \cmidrule(lr){6-6}
	Method & Variant & Anterior & Middle & Posterior &  Foreground \\
	
	\midrule
	
	MLV$^2$-Net & $w_{\text{fg}}=3$ 
    & $\underline{0.626} \pm 0.108$ & $\underline{0.709} \pm 0.092$ & $\underline{0.687} \pm 0.120$ 
    & $\textbf{0.806} \pm 0.030$ 
    
    \\
    MLV$^2$-Net & Oracle
    & $\textbf{0.638} \pm 0.151$ & $\textbf{0.712} \pm 0.094$ & $\textbf{0.688} \pm 0.122$ 
    & - 
    \\
   \midrule

	nnU-Net & 3D Fullres 
    & $0.593 \pm 0.117$ & $0.689 \pm 0.098$ & $0.682 \pm 0.116$
    & $\underline{0.787} \pm 0.046$ 

    \\
	nnU-Net & 2D 
    & $0.618 \pm 0.106$ & $0.707 \pm 0.099$ & $0.683 \pm 0.109$ 
    & $0.760 \pm 0.024$ 

    \\
	\midrule

	UniverSeg & All planes 
    & $0.341 \pm 0.106$ & $0.459 \pm 0.159$ & $0.490 \pm 0.114$ 
    & $0.529 \pm 0.177$ 

    \\

	UniverSeg & Coronal 
    & $0.262 \pm 0.090$ & $0.437 \pm 0.135$ & $0.401 \pm 0.100$ 
    & $0.498 \pm 0.130$ 

    \\
	UniverSeg & Sagittal 
    & $0.312 \pm 0.113$ & $0.411 \pm 0.148$ & $0.413 \pm 0.124$ 
    & $0.427 \pm 0.187$ 

    \\

	UniverSeg & Transverse 
    & $0.275 \pm 0.102$ & $0.318 \pm 0.128$ & $0.366 \pm 0.107$
    & $0.417 \pm 0.113$ 
 
    \\
	\midrule
	SegProp & Optimized  
    & $0.350 \pm 0.137$ & $0.445 \pm 0.104$ & $0.396 \pm 0.110$
    & $0.493 \pm 0.079$ 
 
    \\
	SegProp & Standard 
    & $0.200 \pm 0.078$ & $0.291 \pm 0.072$ & $0.228 \pm 0.080$ 
    & $0.294 \pm 0.065$ 
 
    \\

	\bottomrule
\end{tabular}

}
\end{table*}

\subsection{Experimental setting}
We implemented MLV$^2$-Net into nnU-Net (v2, 3D Fullres), based on Python (v3.11), PyTorch (v2.1), and CUDA (v12.1). 
As there is no reference method for automatic segmentation of MLVs, we compare MLV$^2$-Net to a diverse set of baseline methods. 
Namely, we implemented a registration-based segmentation propagation algorithm~\citep{niftyreg_non_rigid_transform_algorithm} (SegProp) that aggregates all training references through an optimized threshold, UniverSeg~\citep{universeg}, a recent foundation model for medical image segmentation that we adapted for 3D images by fusing overlapping patches from all three image planes, and the standard nnU-Net configurations (2D and 3D Fullres)~\citep{nnunet}. In addition, we implemented an ensemble of separate, rater-specific models (not to be confused with nnU-Net's cross-validation-based ensembling strategy), and we ablate the weighted majority-label voting and the rater-specific labels. 
We ran all methods consistently on a single Nvidia GeForce RTX 3090 graphics card with 24GB VRAM. All experiments were conducted with the data described in the initial paragraph about data and code availability.

\subsection{Inter-rater reliability and rater-based uncertainty}
\label{sec:irr}

\begin{figure}[htpb]
  \centering
  \includegraphics[width=\linewidth]{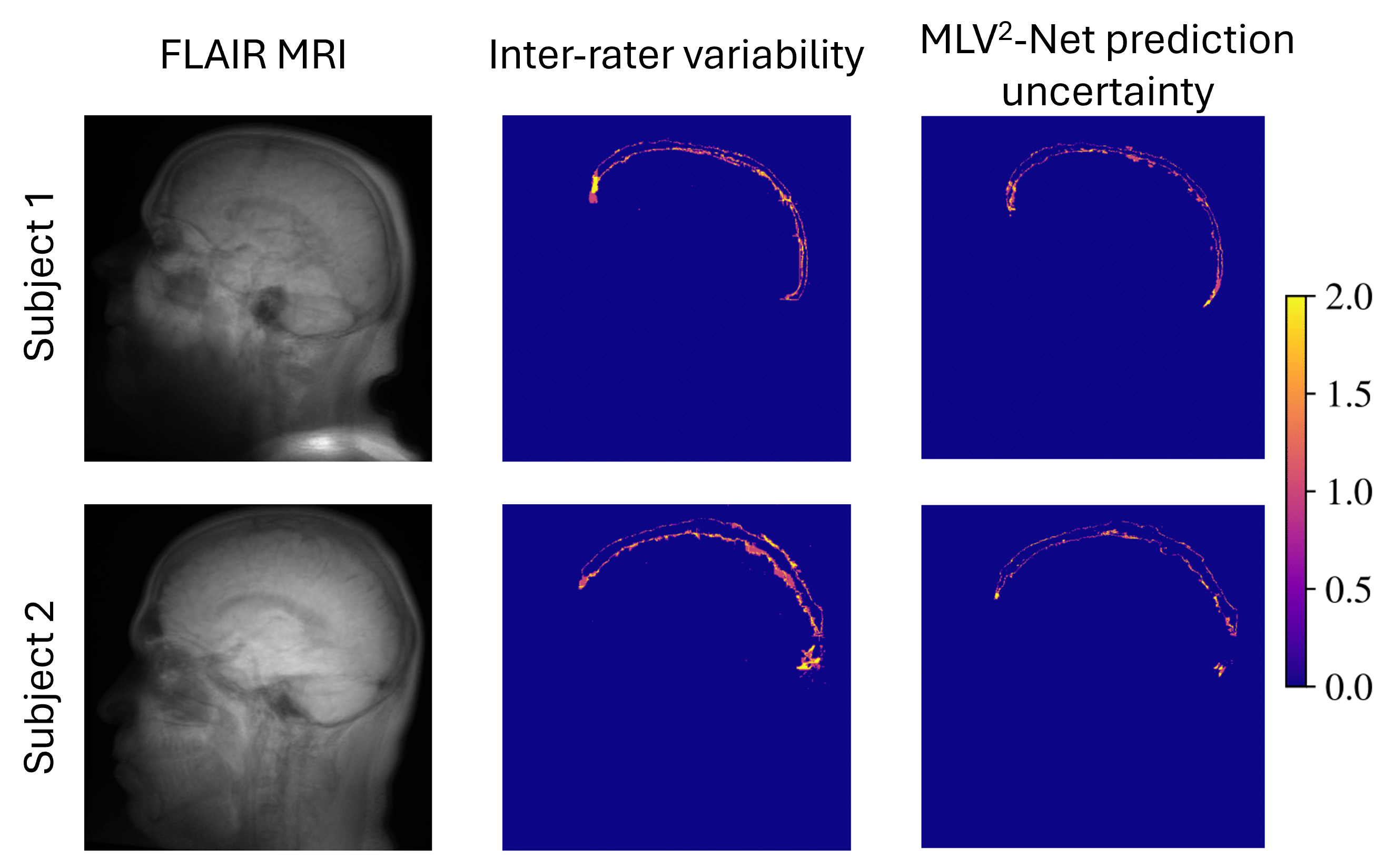}
  \caption{Inter-rater variability between human experts and in MLV$^2$-Net. We plot the disagreement between raters, i.e., a score of one implies that at least one rater disagrees, two means it is a tie, and project voxel-wise values to the sagittal plane. Shown are the two IRR test cases for which individual annotations from all four raters are available.}
  \label{fig:uncertainty}
\end{figure}

\begin{figure*}[t]
  \centering
  \includegraphics[width=0.8\textwidth]{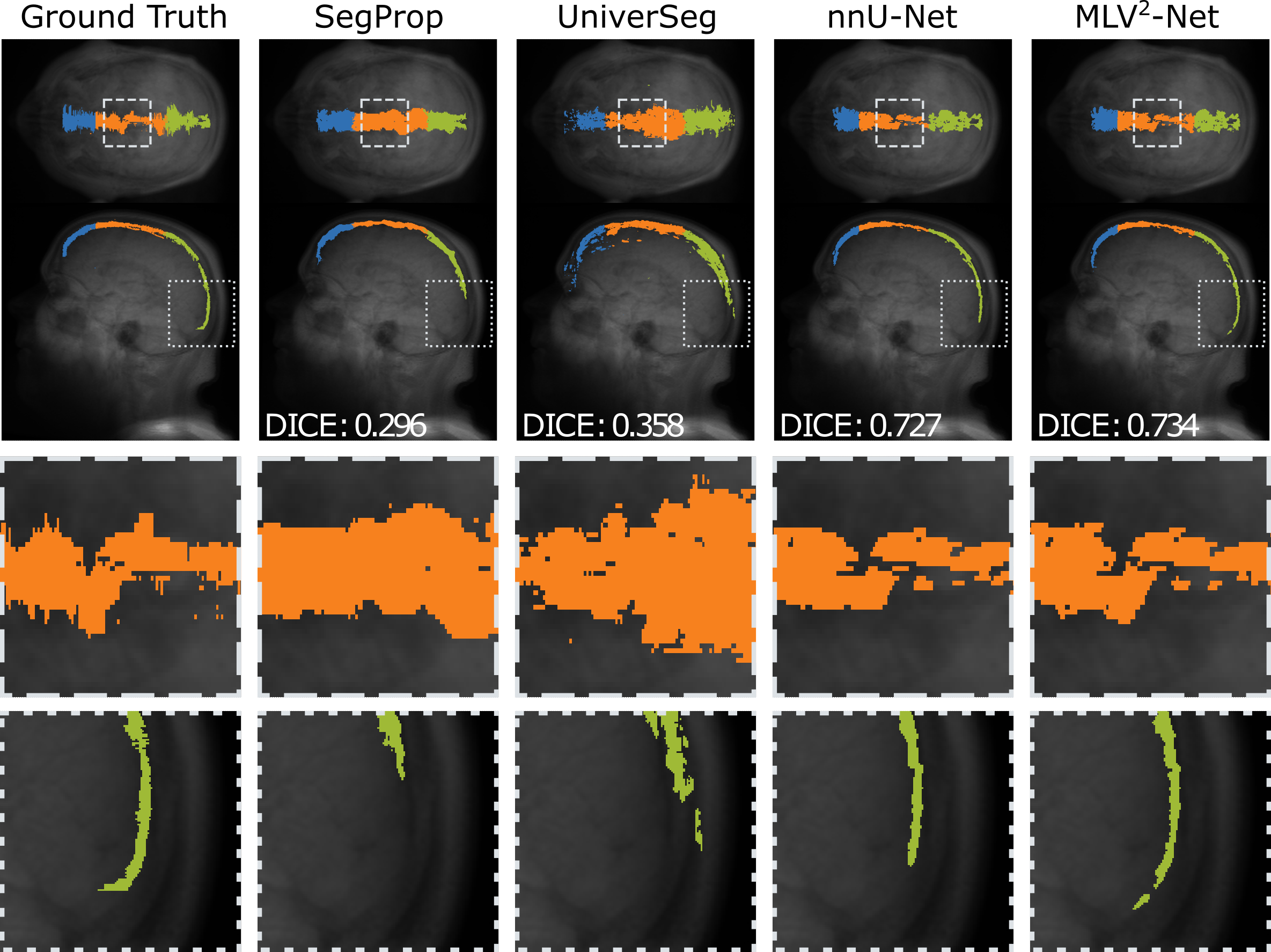}
  \caption{Predictions of all implemented methods based on an image from our consensus test set. The first row shows the orthogonal projections onto axial (top) and sagittal (bottom) planes. We reduced the brightness of the FLAIR image to highlight the segmentation details.} 
  \label{fig:all_best_versions_comparison_example}
\end{figure*}

As a measure of inter-rater reliability (IRR), we compute a Fleiss' kappa score~\citep{Fleiss1971} based on our IRR dataset. This dataset contains an annotation from each rater for each image. Considering all three foreground labels as a single entity, we obtain a Fleiss' kappa of $\kappa=0.73/0.79$ for the two IRR images, respectively.
The ensemble of separate, rater-specific nnU-Net models closely replicates the expert raters' IRR ($\kappa = 0.74/0.80$). With the single-model MLV$^2$-Net, however, we obtain a higher agreement ($\kappa=0.79/0.82$). 

In \Cref{fig:individual-preds}, we show four rater-specific predictions of MLV$^2$-Net for an exemplary scan alongside the annotation of R2. The R2-specific prediction corresponds well to the raters' reference. It can also be seen that conditioning the model on the other raters (R1, R3, R4) yields a reasonable variability in the prediction. 
The vanilla nnU-Net, however, is not capable of modeling this variability and produces only a single prediction.

In \Cref{fig:uncertainty}, we show the rater-based prediction uncertainty of MLV$^2$-Net for the two images in our IRR set and compare it to the voxel-wise inter-rater variability. Qualitatively, the same boundary regions are subject to inter-rater variability and prediction uncertainty. This indicates that the rater-based model uncertainty matches the actual variability in the annotated data locally, thereby supporting the globally computed quantitative IRR results and the qualitative inspection from above.

\subsection{Accuracy and consensus decision-making} \label{sec:consens}

\begin{figure}[t]

    \centering\includegraphics[width=\linewidth]{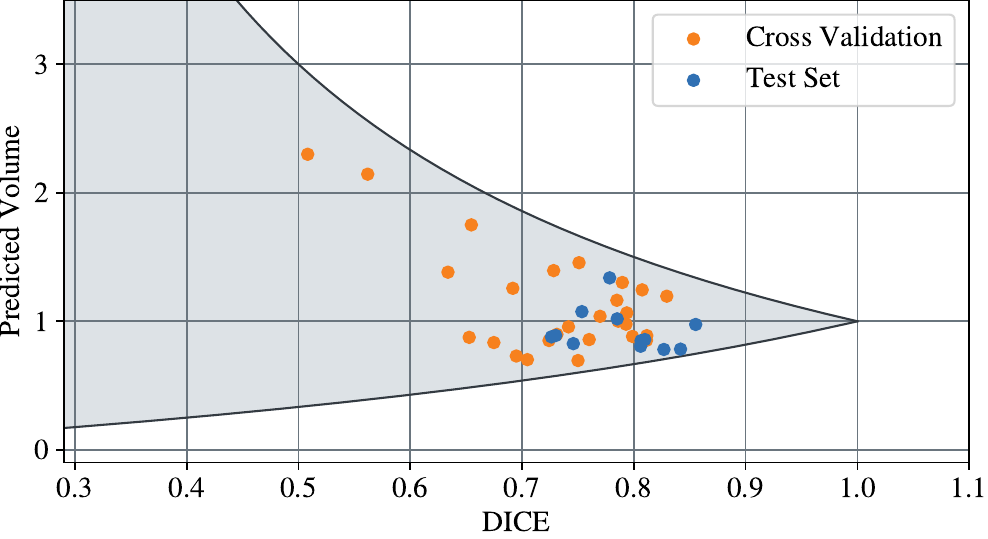}
    \caption{Dice similarity coefficient vs.~relative predicted volume for the cross-validation and annotated test sets. We also show the theoretically derived boundaries on the predicted volume.
    }
    \label{fig:acc-bounds}
\end{figure}

While annotation variability among human raters is natural and unavoidable, most applications demand consistent segmentation. In our held-out consensus test set, we tried to remove the variability as much as possible through all four raters' joint annotation of the images. This is the reference standard for consensus decision-making but, unfortunately, it is only feasible for a few images. In \Cref{tab:accuracy}, we report the average accuracy of all implemented methods on this consensus set and in a 5-fold cross-validation on our training set, separated by anterior, middle, and posterior regions. Qualitative predictions are in \Cref{fig:all_best_versions_comparison_example}. The models evaluated on the consensus set are ensembles of the five cross-validation models.

First, it stands out that MLV$^2$-Net yields the highest accuracy during cross-validation and on the consensus set, followed by the vanilla nnU-Net. Surprisingly, UniverSeg, which was pre-trained on more than 22K medical scans, is not competitive with supervised learning from scratch on our comparably small training set --- even after deliberate optimization. Visually, SegProp also produces reasonable MLV segmentations but is not competitive in terms of quantitative measures.  All models achieve higher Dice scores on the consensus set and sacrifice accuracy in the cross-validation, where the best accuracy is obtained with the respective rater as an input to MLV$^2$-Net (oracle). This is likely due to the higher annotation variability in the training set compared to the consensus set.
Nonetheless, it is noteworthy that MLV$^2$-Net learns to produce highly accurate and consistent segmentations from training data with non-neglectable annotation variability. In \Cref{fig:acc-bounds}, we plot the accuracy of MLV$^2$-Net ($w_{\text{fg}}=3$) and the relative segmented volume in all annotated test samples and from the cross-validation. All results lie within the theoretically derived error bounds. On average, the relative predicted volume can be assumed to be between 0.67 and 1.49 on the consensus set (mean $DSC=0.806$).

\subsection{Ablation study}
\label{sec:ablation}

From the ablation study in \Cref{tab:ablation}, we infer that the most important design choice in MLV$^2$-Net is the weighted majority-label voting. It makes the model more sensitive to foreground voxels than the standard, equally weighted majority vote. Other methodological choices, such as using a single-model approach and rater-specific labels, seem to have only a minor positive effect on segmentation accuracy. 

For our setting with four raters, we can deduce explicit segmentation thresholds in dependence of the foreground weight $w_{\text{fg}}$. In words, $w_{\text{fg}}=3$ means that a voxel is segmented as foreground if no more than two out of the four rater-specific predictions anticipate it to be background (with three background votes, the voxel is predicted as background due to our policy to choose the lower-label index in case of a tie, cf.~\Cref{sec:nnunet_combining_annotator_specific_predictions}). With $w_{\text{fg}}=2$, two votes on the background can only be overruled if the other two votes are on the same MLV sub-label (anterior, middle, posterior), which makes it slightly less foreground-sensitive than $w_{\text{fg}}=3$. Increasing the sensitivity to foreground votes further by setting $w_{\text{fg}}=4$ reduces the performance to the standard majority vote. Thus, we deduce that a moderately increased foreground sensitivity emulates human consensus-finding best based on the given data.

\begin{table}[t]
\setlength{\tabcolsep}{6pt}
    \centering
    \caption{Ablation study of training and consensus-finding strategies. Using our consensus set, we report the mean $\pm$ SD Dice similarity coefficient ($DSC$).}
    \label{tab:ablation}
        {\renewcommand\bfdefault{b}
\begin{tabular}{l c  }
	\toprule
	Configuration & $DSC$  \\
	
	\midrule

    Standard majority vote ($w_{\text{fg}}=1$)
    & $0.790 \pm 0.030$ 
    \\

    Weighted Majority Vote ($w_{\text{fg}}=4$) 
    & $0.790 \pm 0.043$ 
    \\

    No rater-specific labels ($w_{\text{fg}}=3$)
    & $0.800 \pm 0.032$ 
    \\

    Rater-specific models ($w_{\text{fg}}=3$) 
    & $0.804 \pm 0.029$ 
    \\

    Weighted majority vote ($w_{\text{fg}}=2$)
    & $0.805 \pm 0.031$ 
    \\
    
    Weighted majority vote ($w_{\text{fg}}=3$)
    & $0.806 \pm 0.030$ 
    \\

	\bottomrule
\end{tabular}

}    
\end{table}

\subsection{Downstream analysis of MLV volume}
\label{sec:downstream}

\begin{figure}
    \centering\includegraphics[width=0.8\linewidth]{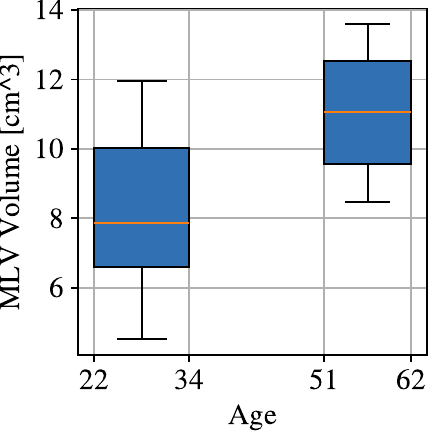}
    \caption{Box plots of the MLV volume predicted by MLV$^2$-Net for two age groups. Lines indicate the median, boxes span the inter-quartile range (IQR), and whiskers extend to $1.5\times$IQR.}
    \label{fig:age-box}
\end{figure}

Finally, we apply our model in a downstream analysis of MLV volume using unlabeled imaging data. Recently, \citet{albayram2022non} found a positive association of age with MLV volume based on manual annotations. Using our MLV$^2$-Net model, we can replicate this finding based on a group of adults ($n=4$, age 51-62) and a larger young reference cohort ($n=18$, age 22-34), see \Cref{fig:age-box}. The difference is significant based on $p<0.05$ (two-sided t-test). This result indirectly confirms the accuracy of our model and proves its applicability for analyzing real-world study data.

\section{Discussion}

\paragraph{MLV segmentation task}
The task of MLV segmentation is new and challenging due to the ramified structure, the thin diameter, and the high inter-rater variability among experts on the voxel level. The difficulty of segmenting these structures is reflected in the high inter-rater variability in the expert-annotated data (Fleiss' kappa of $\kappa=0.73/0.79$), cf.~\Cref{sec:irr}. According to \citet{Landis1977}, this corresponds to a \emph{substantial agreement} ($0.6 < \kappa \leq 0.8$), which is inferior to \emph{perfect agreement} ($\kappa > 0.8$). 
Nonetheless, we achieved a high accuracy of $DSC=0.806$ on our consensus test set. 
As no established baselines exist for this task, we tried to cover various methods ranging from segmentation propagation over supervised learning to recent foundation models, cf.~\Cref{sec:consens}. Yet, future research should investigate and compare alternative model architectures and training paradigms to draw a more complete picture of the task at hand.

\paragraph{Dataset size}
We are aware that the number of annotated scans ($n=33$) used in this study is comparably small, especially when compared to recent segmentation datasets with annotations of thousands of anatomies~\citep{Wasserthal2023}. Yet, datasets with around 30 annotated scans are not uncommon for 3D medical image segmentation~\citep{Antonelli2022}. In fact, creating much larger segmentation datasets manually for MLV structures is impossible due to their complex shape, the low contrast even in FLAIR imaging, and the required high resolution of 0.5mm in sagittal and vertical axes. In our experiments, we tried to account for the small dataset size by tuning hyperparameters based on extensive cross-validation on the training set. Moreover, we indirectly assessed our model's performance on $n=22$ unannotated scans (cf.~\Cref{sec:downstream}) by replicating known age-related associations with MLV volume. Finally, we put particular effort into assessing the inter-rater reliability (cf.~\Cref{sec:irr}) and created a consensus test set to ensure the used annotations are of high quality.

\paragraph{Foreground bias in ensemble decision-making}
In our ablation study in \Cref{sec:ablation}, we found a foreground weight of $w_{\text{fg}}=3$ to work best for the given MLV datasets. This essentially creates a bias toward foreground labels, which seems to mimic human consensus decision-making to a certain degree in our case. However, it is unclear how this observation generalizes to other data, structures, and rater groups. Albeit an analysis of this relation is out of scope for this paper, it could be an interesting starting point for follow-up research to investigate the observed foreground annotation bias and its implications.

\section{Conclusion}
In summary, we presented the first automatic method for MLV segmentation from 3D FLAIR imaging. 
Our model, MLV$^2$-Net, outperformed state-of-the-art baselines by embracing the styles of all annotators involved in the creation of the training set. 
In contrast to most segmentation methods, MLV$^2$-Net provides a rater-based uncertainty estimation. Together with the derived theoretical bounds on the segmented volume, we expect MLV$^2$-Net to be a valuable tool for clinical researchers that study the glymphatic system. Yet, the technical contributions and code are generic and could be beneficial for other applications as well.

\acks{
This research was partially supported by the German Research Foundation.
}

\bibliography{references}

\begin{thebibliography}{23}
\providecommand{\natexlab}[1]{#1}
\providecommand{\url}[1]{\texttt{#1}}
\expandafter\ifx\csname urlstyle\endcsname\relax
  \providecommand{\doi}[1]{doi: #1}\else
  \providecommand{\doi}{doi: \begingroup \urlstyle{rm}\Url}\fi

\bibitem[Albayram et~al.(2022)Albayram, Smith, Tufan, Tuna, Bostanc{\i}kl{\i}o{\u{g}}lu, Zile, and Albayram]{albayram2022non}
Mehmet~Sait Albayram, Garrett Smith, Fatih Tufan, Ibrahim~Sacit Tuna, Mehmet Bostanc{\i}kl{\i}o{\u{g}}lu, Michael Zile, and Onder Albayram.
\newblock Non-invasive mr imaging of human brain lymphatic networks with connections to cervical lymph nodes.
\newblock \emph{Nature communications}, 13\penalty0 (1):\penalty0 203, 2022.

\bibitem[Antonelli et~al.(2022)Antonelli, Reinke, Bakas, Farahani, Kopp-Schneider, Landman, Litjens, Menze, Ronneberger, Summers, van Ginneken, Bilello, Bilic, Christ, Do, Gollub, Heckers, Huisman, Jarnagin, McHugo, Napel, Pernicka, Rhode, Tobon-Gomez, Vorontsov, Meakin, Ourselin, Wiesenfarth, Arbeláez, Bae, Chen, Daza, Feng, He, Isensee, Ji, Jia, Kim, Maier-Hein, Merhof, Pai, Park, Perslev, Rezaiifar, Rippel, Sarasua, Shen, Son, Wachinger, Wang, Wang, Xia, Xu, Xu, Zheng, Simpson, Maier-Hein, and Cardoso]{Antonelli2022}
Michela Antonelli, Annika Reinke, Spyridon Bakas, Keyvan Farahani, Annette Kopp-Schneider, Bennett~A. Landman, Geert Litjens, Bjoern Menze, Olaf Ronneberger, Ronald~M. Summers, Bram van Ginneken, Michel Bilello, Patrick Bilic, Patrick~F. Christ, Richard K.~G. Do, Marc~J. Gollub, Stephan~H. Heckers, Henkjan Huisman, William~R. Jarnagin, Maureen~K. McHugo, Sandy Napel, Jennifer S.~Golia Pernicka, Kawal Rhode, Catalina Tobon-Gomez, Eugene Vorontsov, James~A. Meakin, Sebastien Ourselin, Manuel Wiesenfarth, Pablo Arbeláez, Byeonguk Bae, Sihong Chen, Laura Daza, Jianjiang Feng, Baochun He, Fabian Isensee, Yuanfeng Ji, Fucang Jia, Ildoo Kim, Klaus Maier-Hein, Dorit Merhof, Akshay Pai, Beomhee Park, Mathias Perslev, Ramin Rezaiifar, Oliver Rippel, Ignacio Sarasua, Wei Shen, Jaemin Son, Christian Wachinger, Liansheng Wang, Yan Wang, Yingda Xia, Daguang Xu, Zhanwei Xu, Yefeng Zheng, Amber~L. Simpson, Lena Maier-Hein, and M.~Jorge Cardoso.
\newblock The medical segmentation decathlon.
\newblock \emph{Nature Communications}, 13\penalty0 (1), July 2022.

\bibitem[Butoi et~al.(2023)Butoi, Ortiz, Ma, Sabuncu, Guttag, and Dalca]{universeg}
Victor~Ion Butoi, Jose Javier~Gonzalez Ortiz, Tianyu Ma, Mert~R. Sabuncu, John Guttag, and Adrian~V. Dalca.
\newblock Universeg: Universal medical image segmentation.
\newblock \emph{International Conference on Computer Vision}, 2023.

\bibitem[{Del Guerra} et~al.(2018){Del Guerra}, Ahmad, Avram, Belcari, Berneking, Biagi, Bisogni, Brandl, Cabello, Camarlinghi, Cerello, Choi, Coli, Colpo, Fleury, Gagliardi, Giraudo, Heekeren, Kawohl, Kostou, Lefaucheur, Lerche, Loudos, Morrocchi, Muller, Mustafa, Neuner, Papadimitroulas, Pennazio, Rajkumar, Brambilla, Rivoire, Kops, Scheins, Schimpf, Shah, Sorg, Sportelli, Tosetti, Trinchero, Wyss, and Ziegler]{trimage}
Alberto {Del Guerra}, Salleh Ahmad, Mihai Avram, Nicola Belcari, Arne Berneking, Laura Biagi, Maria~Giuseppina Bisogni, Felix Brandl, Jorge Cabello, Niccolò Camarlinghi, Piergiorgio Cerello, Chang-Hoon Choi, Silvia Coli, Sabrina Colpo, Julien Fleury, Vito Gagliardi, Giuseppe Giraudo, Karsten Heekeren, Wolfram Kawohl, Theodora Kostou, Jean-Luc Lefaucheur, Christoph Lerche, George Loudos, Matteo Morrocchi, Julien Muller, Mona Mustafa, Irene Neuner, Panagiotis Papadimitroulas, Francesco Pennazio, Ravichandran Rajkumar, Cláudia~Régio Brambilla, Julien Rivoire, Elena~Rota Kops, Jürgen Scheins, Rémy Schimpf, N.~Jon Shah, Christian Sorg, Giancarlo Sportelli, Michela Tosetti, Riccardo Trinchero, Christine Wyss, and Sibylle Ziegler.
\newblock Trimage: A dedicated trimodality (pet/mr/eeg) imaging tool for schizophrenia.
\newblock \emph{European Psychiatry}, 50:\penalty0 7--20, 2018.

\bibitem[Ding et~al.(2021)Ding, Wang, Xia, Liu, Tian, Fu, Chen, Qin, Wang, Xiang, Zhang, Cao, Wang, Li, Wu, Tang, Ma, Teng, and Wang]{Ding2021}
Xue-Bing Ding, Xin-Xin Wang, Dan-Hao Xia, Han Liu, Hai-Yan Tian, Yu~Fu, Yong-Kang Chen, Chi Qin, Jiu-Qi Wang, Zhi Xiang, Zhong-Xian Zhang, Qin-Chen Cao, Wei Wang, Jia-Yi Li, Erxi Wu, Bei-Sha Tang, Ming-Ming Ma, Jun-Fang Teng, and Xue-Jing Wang.
\newblock Impaired meningeal lymphatic drainage in patients with idiopathic parkinson’s disease.
\newblock \emph{Nature Medicine}, 27\penalty0 (3):\penalty0 411–418, January 2021.

\bibitem[Fleiss(1971)]{Fleiss1971}
Joseph~L. Fleiss.
\newblock Measuring nominal scale agreement among many raters.
\newblock \emph{Psychological Bulletin}, 76\penalty0 (5):\penalty0 378–382, November 1971.

\bibitem[Gal and Ghahramani(2016)]{gal16dropout}
Yarin Gal and Zoubin Ghahramani.
\newblock Dropout as a bayesian approximation: Representing model uncertainty in deep learning.
\newblock In Maria~Florina Balcan and Kilian~Q. Weinberger, editors, \emph{Proceedings of The 33rd International Conference on Machine Learning}, volume~48 of \emph{Proceedings of Machine Learning Research}, pages 1050--1059, New York, New York, USA, 20--22 Jun 2016. PMLR.

\bibitem[Goodman et~al.(2018)Goodman, Adham, Woltjer, Lund, and Iliff]{Goodman2018}
James~R. Goodman, Zachariah~O. Adham, Randall~L. Woltjer, Amanda~W. Lund, and Jeffrey~J. Iliff.
\newblock Characterization of dural sinus-associated lymphatic vasculature in human alzheimer’s dementia subjects.
\newblock \emph{Brain, Behavior, and Immunity}, 73:\penalty0 34–40, October 2018.

\bibitem[Guo et~al.(2024)Guo, Lin, Yang, Yu, Cheng, and Yan]{Guo2024}
Xiayu Guo, Xian Lin, Xin Yang, Li~Yu, Kwang-Ting Cheng, and Zengqiang Yan.
\newblock Uctnet: Uncertainty-guided cnn-transformer hybrid networks for medical image segmentation.
\newblock \emph{Pattern Recognition}, 152:\penalty0 110491, August 2024.

\bibitem[Hatamizadeh et~al.(2022)Hatamizadeh, Tang, Nath, Yang, Myronenko, Landman, Roth, and Xu]{hatamizadeh2022unetr}
Ali Hatamizadeh, Yucheng Tang, Vishwesh Nath, Dong Yang, Andriy Myronenko, Bennett Landman, Holger~R Roth, and Daguang Xu.
\newblock Unetr: Transformers for 3d medical image segmentation.
\newblock In \emph{Proceedings of the IEEE/CVF winter conference on applications of computer vision}, pages 574--584, 2022.

\bibitem[Iliff et~al.(2012)Iliff, Wang, Liao, Plogg, Peng, Gundersen, Benveniste, Vates, Deane, Goldman, Nagelhus, and Nedergaard]{Iliff2012}
Jeffrey~J. Iliff, Minghuan Wang, Yonghong Liao, Benjamin~A. Plogg, Weiguo Peng, Georg~A. Gundersen, Helene Benveniste, G.~Edward Vates, Rashid Deane, Steven~A. Goldman, Erlend~A. Nagelhus, and Maiken Nedergaard.
\newblock A paravascular pathway facilitates csf flow through the brain parenchyma and the clearance of interstitial solutes, including amyloid $\beta$.
\newblock \emph{Science Translational Medicine}, 4\penalty0 (147), August 2012.

\bibitem[Isensee et~al.(2021)Isensee, Jaeger, Kohl, Petersen, and Maier-Hein]{nnunet}
Fabian Isensee, Paul~F. Jaeger, Simon A.~A. Kohl, Jens Petersen, and Klaus~H. Maier-Hein.
\newblock nnu-net: a self-configuring method for deep learning-based biomedical image segmentation.
\newblock \emph{Nature Methods}, 18\penalty0 (2):\penalty0 203--211, 2 2021.

\bibitem[Isensee et~al.(2024)Isensee, Wald, Ulrich, Baumgartner, Roy, Maier-Hein, and Jäger]{isensee2024nnunet-revisited}
Fabian Isensee, Tassilo Wald, Constantin Ulrich, Michael Baumgartner, Saikat Roy, Klaus Maier-Hein, and Paul~F. Jäger.
\newblock { nnU-Net Revisited: A Call for Rigorous Validation in 3D Medical Image Segmentation }.
\newblock In \emph{proceedings of Medical Image Computing and Computer Assisted Intervention -- MICCAI 2024}, volume LNCS 15009. Springer Nature Switzerland, October 2024.

\bibitem[Kohl et~al.(2018)Kohl, Romera-Paredes, Meyer, Fauw, Ledsam, Maier-Hein, Eslami, Rezende, and Ronneberger]{kohl2018-probabilistic-unet}
Simon A.~A. Kohl, Bernardino Romera-Paredes, Clemens Meyer, Jeffrey~De Fauw, Joseph~R. Ledsam, Klaus~H. Maier-Hein, S.~M.~Ali Eslami, Danilo~Jimenez Rezende, and Olaf Ronneberger.
\newblock A probabilistic u-net for segmentation of ambiguous images.
\newblock In \emph{Proceedings of the 32nd International Conference on Neural Information Processing Systems}, NeurIPS'18, page 6965–6975, Red Hook, NY, USA, 2018. Curran Associates Inc.

\bibitem[Landis and Koch(1977)]{Landis1977}
J.~Richard Landis and Gary~G. Koch.
\newblock The measurement of observer agreement for categorical data.
\newblock \emph{Biometrics}, 33\penalty0 (1):\penalty0 159, March 1977.

\bibitem[Louveau et~al.(2015)Louveau, Smirnov, Keyes, Eccles, Rouhani, Peske, Derecki, Castle, Mandell, Lee, Harris, and Kipnis]{Louveau2015}
Antoine Louveau, Igor Smirnov, Timothy~J. Keyes, Jacob~D. Eccles, Sherin~J. Rouhani, J.~David Peske, Noel~C. Derecki, David Castle, James~W. Mandell, Kevin~S. Lee, Tajie~H. Harris, and Jonathan Kipnis.
\newblock Structural and functional features of central nervous system lymphatic vessels.
\newblock \emph{Nature}, 523\penalty0 (7560):\penalty0 337–341, June 2015.

\bibitem[Louveau et~al.(2018)Louveau, Herz, Alme, Salvador, Dong, Viar, Herod, Knopp, Setliff, Lupi, Da~Mesquita, Frost, Gaultier, Harris, Cao, Hu, Lukens, Smirnov, Overall, Oliver, and Kipnis]{Louveau2018}
Antoine Louveau, Jasmin Herz, Maria~Nordheim Alme, Andrea~Francesca Salvador, Michael~Q. Dong, Kenneth~E. Viar, S.~Grace Herod, James Knopp, Joshua~C. Setliff, Alexander~L. Lupi, Sandro Da~Mesquita, Elizabeth~L. Frost, Alban Gaultier, Tajie~H. Harris, Rui Cao, Song Hu, John~R. Lukens, Igor Smirnov, Christopher~C. Overall, Guillermo Oliver, and Jonathan Kipnis.
\newblock Cns lymphatic drainage and neuroinflammation are regulated by meningeal lymphatic vasculature.
\newblock \emph{Nature Neuroscience}, 21\penalty0 (10):\penalty0 1380–1391, September 2018.

\bibitem[Mirikharaji et~al.(2021)Mirikharaji, Abhishek, Izadi, and Hamarneh]{deep_learning_esnambles_from_multiple_annotations}
Zahra Mirikharaji, Kumar Abhishek, Saeed Izadi, and Ghassan Hamarneh.
\newblock D-lema: Deep learning ensembles from multiple annotations - application to skin lesion segmentation.
\newblock In \emph{2021 IEEE/CVF Conference on Computer Vision and Pattern Recognition Workshops (CVPRW)}, pages 1837--1846, 2021.

\bibitem[Modat et~al.(2009)Modat, Ridgway, Taylor, Lehmann, Barnes, Hawkes, Fox, and Ourselin]{niftyreg_non_rigid_transform_algorithm}
Marc Modat, Gerard~R Ridgway, Zeike~A Taylor, Manja Lehmann, Josephine Barnes, David~J Hawkes, Nick~C Fox, and S{\'e}bastien Ourselin.
\newblock Fast free-form deformation using graphics processing units.
\newblock \emph{Comput Methods Programs Biomed}, 98\penalty0 (3):\penalty0 278--284, October 2009.

\bibitem[Ronneberger et~al.(2015)Ronneberger, Fischer, and Brox]{ronnebergerUNetConvolutionalNetworks2015}
Olaf Ronneberger, Philipp Fischer, and Thomas Brox.
\newblock U-{{Net}}: {{Convolutional Networks}} for {{Biomedical Image Segmentation}}.
\newblock In Nassir Navab, Joachim Hornegger, William~M. Wells, and Alejandro~F. Frangi, editors, \emph{Medical {{Image Computing}} and {{Computer-Assisted Intervention}} – {{MICCAI}} 2015}, volume 9351, pages 234--241. {Springer International Publishing}, 2015.

\bibitem[Warfield et~al.(2004)Warfield, Zou, and Wells]{Warfield2004}
S.K. Warfield, K.H. Zou, and W.M. Wells.
\newblock Simultaneous truth and performance level estimation (staple): An algorithm for the validation of image segmentation.
\newblock \emph{IEEE Transactions on Medical Imaging}, 23\penalty0 (7):\penalty0 903–921, July 2004.

\bibitem[Wasserthal et~al.(2023)Wasserthal, Breit, Meyer, Pradella, Hinck, Sauter, Heye, Boll, Cyriac, Yang, Bach, and Segeroth]{Wasserthal2023}
Jakob Wasserthal, Hanns-Christian Breit, Manfred~T. Meyer, Maurice Pradella, Daniel Hinck, Alexander~W. Sauter, Tobias Heye, Daniel~T. Boll, Joshy Cyriac, Shan Yang, Michael Bach, and Martin Segeroth.
\newblock Totalsegmentator: Robust segmentation of 104 anatomic structures in ct images.
\newblock \emph{Radiology: Artificial Intelligence}, 5\penalty0 (5), September 2023.

\bibitem[Zhang et~al.(2023)Zhang, Tanno, Xu, Huang, Bronik, Jin, Jacob, Zheng, Shao, Ciccarelli, Barkhof, and Alexander]{learning_from_multiple_annotators_for_medical_image_segmentation}
Le~Zhang, Ryutaro Tanno, Moucheng Xu, Yawen Huang, Kevin Bronik, Chen Jin, Joseph Jacob, Yefeng Zheng, Ling Shao, Olga Ciccarelli, Frederik Barkhof, and Daniel~C. Alexander.
\newblock Learning from multiple annotators for medical image segmentation.
\newblock \emph{Pattern Recognition}, 138:\penalty0 109400, 2023.

\end{thebibliography}

\end{document}